\documentclass[runningheads]{llncs}
 
\usepackage{eccv}

\usepackage{eccvabbrv}

\usepackage{graphicx}
\usepackage{booktabs}
\usepackage{multirow}
\usepackage{algorithm}
\usepackage{algpseudocode}
\usepackage{colortbl}
\usepackage{threeparttable}
\usepackage{makecell}

\usepackage{hyperref}

\begin{document}
\title{Cross-Domain Semantic Segmentation on Inconsistent Taxonomy using VLMs}

\titlerunning{Cross-Domain Semantic Segmentation on Inconsistent Tax. using VLMs}

\author{Jeongkee Lim\inst{1} \and
Yusung Kim\inst{2}}

\authorrunning{J. Lim et al.}

\institute{Department of Electrical and Computer Engineering, Sungkyunkwan University \and Department of Computer Science and Engineering, Sungkyunkwan University\\
\email{skjkee58@g.skku.edu}, \email{yskim525@skku.edu}}

\maketitle

\begin{abstract}
    The challenge of semantic segmentation in Unsupervised Domain Adaptation (UDA) emerges not only from domain shifts between source and target images but also from discrepancies in class taxonomies across domains. Traditional UDA research assumes consistent taxonomy between the source and target domains, thereby limiting their ability to recognize and adapt to the taxonomy of the target domain. This paper introduces a novel approach, Cross-Domain Semantic Segmentation on Inconsistent Taxonomy using Vision Language Models (CSI), which effectively performs domain-adaptive semantic segmentation even in situations of source-target class mismatches. CSI leverages the semantic generalization potential of Visual Language Models (VLMs) to create synergy with previous UDA methods.
    It leverages segment reasoning obtained through traditional UDA methods, combined with the rich semantic knowledge embedded in VLMs, to relabel new classes in the target domain. This approach allows for effective adaptation to extended taxonomies without requiring any ground truth label for the target domain. Our method has shown to be effective across various benchmarks in situations of inconsistent taxonomy settings (coarse-to-fine taxonomy and open taxonomy) and demonstrates consistent synergy effects when integrated with previous state-of-the-art UDA methods. The implementation is available at \url{https://github.com/jkee58/CSI}.
\end{abstract}

\begin{figure}[tb]
  \centering
  \includegraphics[width=0.7\linewidth]{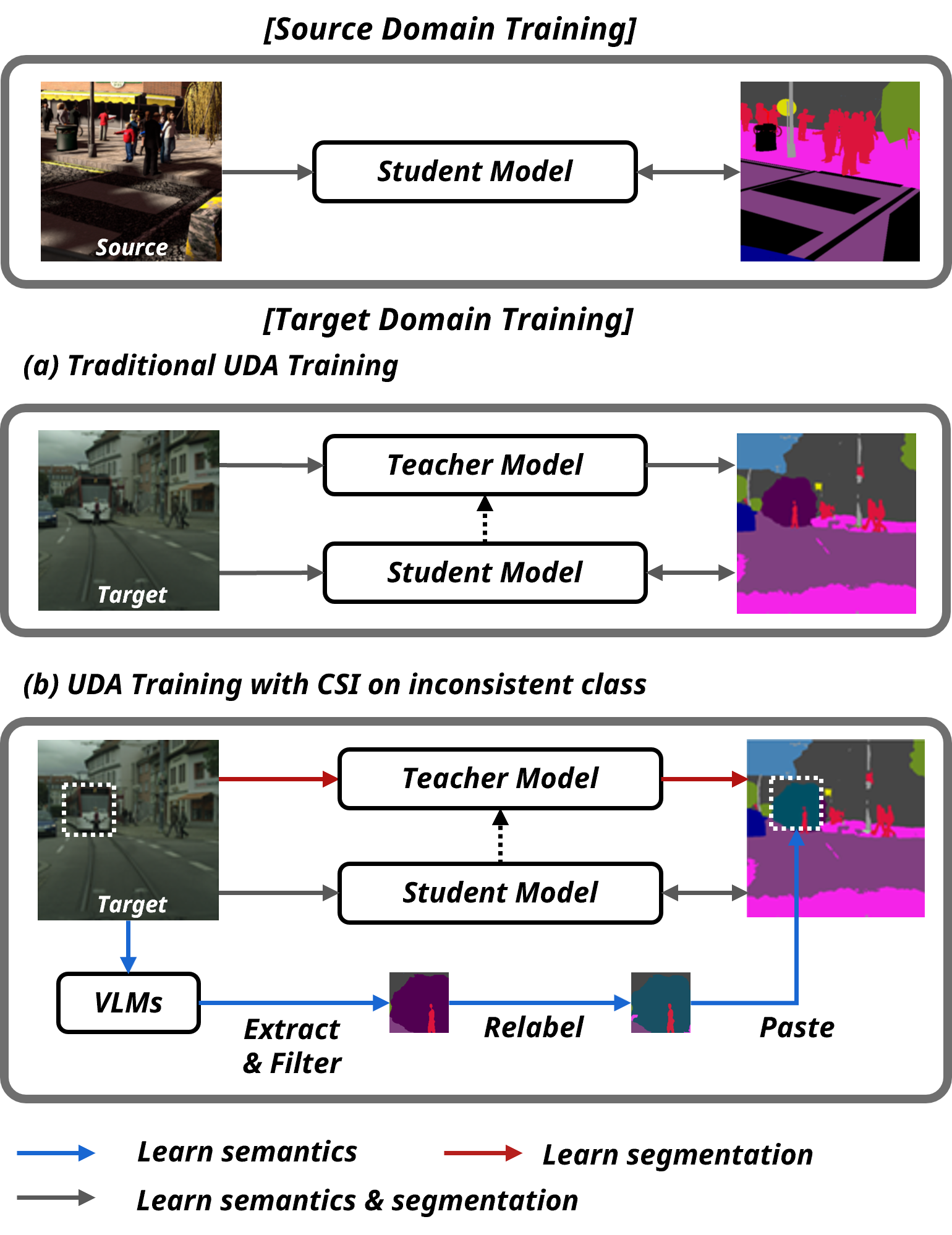}
  \caption{UDA consists of supervised learning on the source domain and unsupervised learning on the target domain. (a) The traditional UDA model learns all reasoning about the target domain from the teacher model. (b) Our CSI leverages VLMs to learn semantic information that does not exist in the target domain and combines it with known segment reasoning.}
  \label{fig:brief}
\end{figure}

\section{Introduction}
Semantic segmentation plays a crucial role in various computer vision applications, from autonomous driving to medical image analysis. The essence of this task is to classify each pixel in an image into a predefined set of categories, allowing to understand and interpret the visual world in detail. While semantic segmentation can be significantly utilized across diverse application services, manually creating pixel-level ground truth labels for training is a challenging and costly endeavor. Unsupervised Domain Adaptation (UDA) has emerged as a powerful tool for resolving the domain distribution difference by training on a source domain (e.g., synthetic images) with labeled data and then applying it to a target domain (e.g., real-world images) where labeled data is unavailable. However, traditional UDA approaches presuppose a consistent taxonomy across domains, limiting their applicability in real scenarios where the class taxonomies between the source and target domains often differ due to different scenarios or requirements.

The class taxonomy inconsistency between source and target domains discussed in this paper can be divided into two types. One is open taxonomy, where classes that do not exist in the source domain appear in the target domain. The second is coarse-to-fine taxonomy, where a single class in the source domain is split into multiple classes in the target domain (e.g., \textit{car} to \textit{car} and \textit{truck}). Existing Open-Set Domain Adaptation research \cite{saito2018open, panareda2017open, you2019universal, yu2023open} has primarily focused on recognizing newly appearing unseen classes in the target domain as an unknown class, mostly limited to the Image Classification task. As for Domain Adaptive Semantic Segmentation research, TACS \cite{gong2022tacs} was the first to challenge the open, coarse-to-fine, and implicitly overlapping taxonomy. However, it has the limitation of requiring a small amount of labeled data for the target domain.

The recent advancements in Vision Language Models (VLMs) have opened new possibilities for semantic understanding beyond domain-specific learning. This paper introduces a novel approach, Cross-Domain Semantic Segmentation on Inconsistent using VLMs (CSI), which leverages the semantic generalization capabilities of VLMs to address the class taxonomy inconsistency problem in UDA. Our new methodology proposes a solution that can adapt to changed class taxonomies in the target domain without using any label data, tackling both open and coarse-to-fine taxonomy issues. By integrating spatial segmentation reasoning inferred from traditional UDA methods with the rich semantic knowledge embedded in VLMs, CSI can more flexibly and effectively solve UDA problems with taxonomy inconsistencies.

As shown in Fig. \ref{fig:brief}, this paper proposes a simple method for performing zero-shot relabeling inconsistent taxonomy using OWL-ViT \cite{minderer2022simple} and CLIP \cite{radford2021learning}, based on segmentation pseudo-labels inferred from traditional teacher network-based UDA approaches. Through extensive experimentation on various benchmarks exhibiting class taxonomy discrepancies, we demonstrate the effectiveness of our approach and its compatibility with various existing UDA methods. This research is the first to show that zero-shot relabeling can address the class taxonomy inconsistency problem in UDA, not only enabling adaptation to classes inconsistent with the source domain but also improving segmentation adaptation performance for classes common to both source and target domains.

\section{Related Work}
\subsection{Semantic Segmentation with Domain Adaptation}
\subsubsection{Semantic Segmentation}
In computer vision, semantic segmentation is a perception task that is widely applied in several practical applications such as autonomous driving, robotics, and medical imaging. Thanks to the development of computer hardware and the surge of deep learning, semantic segmentation has made remarkable improvements in recent years. In particular, inspired by the success of transformers in image classification \cite{dosovitskiy2020image}, many studies adapted it to semantic segmentation \cite{xie2021segformer, liu2021swin, zheng2021rethinking} achieving remarkable performance. In addition to research on improving the accuracy of semantic segmentation, there is also research focused on balancing inference speed and model precision \cite{yu2018bisenet, fan2022mlfnet}.

To train these models, a large amount of high-quality data annotated with a predefined class is required. Humans are the most accurate way to create high-quality annotations for real-world images at present. However, annotating images by humans is very time-consuming and labor-intensive. To address these issues, several frameworks have been proposed such as Unsupervised Domain Adaptation.

\subsubsection{Unsupervised Domain Adaptation}
Unsupervised Domain Adaptation (UDA) is a framework that adapts networks for the source domain with several labels to the target domain without labels. The massive labeled data in the source domain is mainly generated in a virtual environment or uses a pre-collected dataset. Therefore, UDA can be an alternative to supervised learning for semantic segmentation because it does not require a large amount of annotated images by humans.

According to recent studies, UDA can be divided into two groups: adversarial training and self-training. Adversarial training \cite{ganin2016domain, goodfellow2014generative} aims to align the distribution between the source domain and the target domain in input \cite{gong2019dlow, hoffman2018cycada}, feature \cite{ganin2016domain, hoffman2016fcns, long2018conditional}, and output space \cite{luo2021category, saito2018maximum, tsai2018learning, vu2019advent}. However, adversarial training has problems with mode collapse and training instability. Unlike adversarial training, self-training is training the model with pseudo labels \cite{pseudo2013simple} for the target domain. The pseudo labels can be obtained based on confidence \cite{mei2020instance, zhang2018collaborative, zou2018unsupervised} or prototypes \cite{pan2019transferrable, zhang2021prototypical, zhang2019category}. In self-training, improving the quality of the pseudo label is important to increase the robustness of the model. So, various methods aim to ensure consistency by leveraging data augmentations \cite{araslanov2021self, choi2019self, french2017self, melas2021pixmatch, prabhu2021sentry} or domain-mixup \cite{gao2021dsp, hoyer2022daformer, hoyer2022hrda, hoyer2023improving, liu2021bapa, tranheden2021dacs, zhou2022context, kim2023bidirectional}. Recently, various models, including the SOTA model, have used the self-training method.

However, these traditional UDA methods are only applicable in a consistent taxonomy setting where the source and target domains are the same. This is not a suitable setting for solving practical problems. Some studies \cite{saito2018open, panareda2017open, you2019universal} propose methods to recognize unseen classes, which are classes only in the target domain. In TACS \cite{gong2022tacs}, inconsistent taxonomies are categorized into open, coarse-to-fine, and implicitly overlapping taxonomies. We go beyond these studies and propose a method for UDA in the mixed type of inconsistent taxonomies (open and coarse-to-fine) without using the GT label of the target domain.

\subsection{Vision Language Models (VLMs)}

Several large image language models, including CLIP \cite{radford2021learning}, utilize large image-text datasets collected from the web to learn the joint embedding space between images and text. These models have good generalization capabilities because they are trained on a large amount of diverse datasets. In addition, they utilize both images and text, which allows for the application of open vocabularies, whereas previous models were limited to fixed label spaces. This has significantly improved the performance of many zero-shot tasks, such as object detection and semantic segmentation.

Much recent research has aimed to extend the open vocabulary capabilities of large image language models to tasks such as object detection, among other zero-shot tasks. Object detection using open vocabularies has made significant progress by combining image-text models such as CLIP with traditional object detection models. Recent proposals include adding a frozen image language model encoder \cite{kuo2022open} to the detection head, utilizing a fine-tuned image language model \cite{minderer2022simple} or an open classification model for cropped image regions \cite{gu2021open}. 

In addition to object detection, various zero-shot techniques utilizing CLIP have been proposed for semantic segmentation. MaskCLIP \cite{zhou2022extract} showed that CLIP's image encoder allows segmentation without additional training. However, since it does not utilize segmentation labels, the predicted results are noisy and perform poorly, making it difficult to use for unsupervised domain adaptation.

\begin{figure}[tb]
  \centering
  \includegraphics[width=1.0\linewidth]{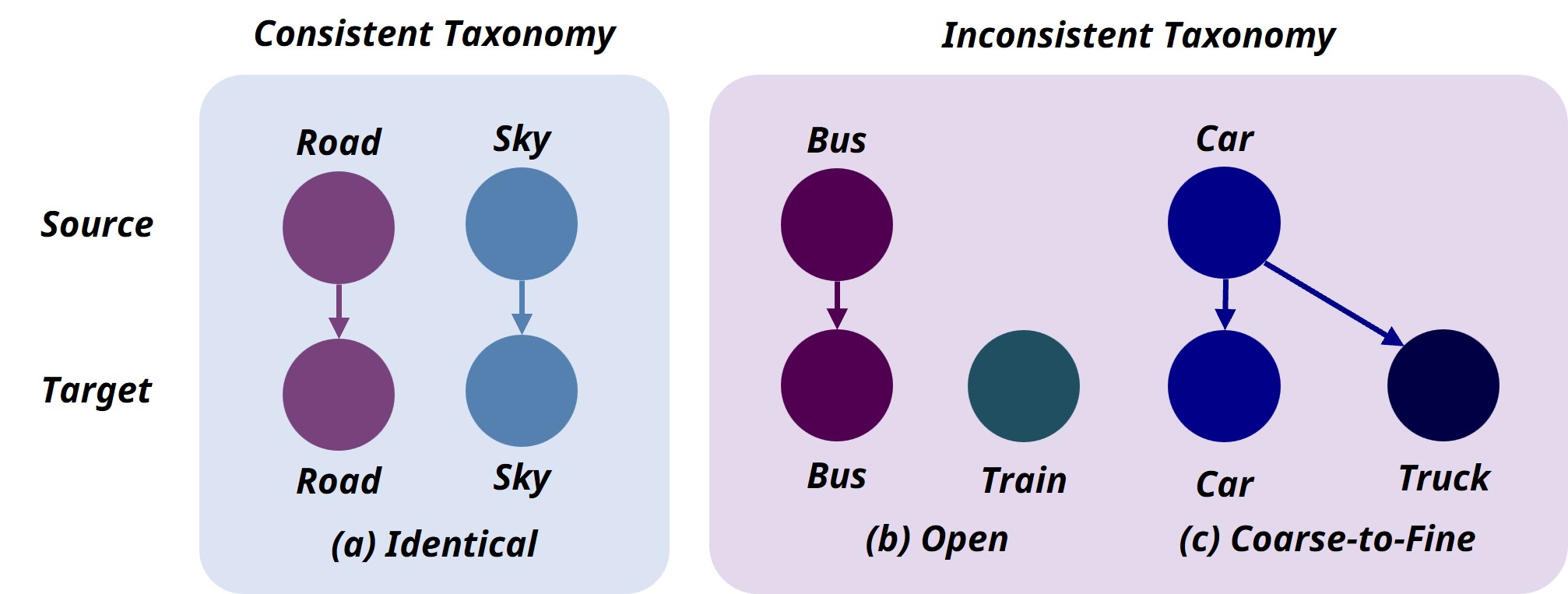}
  \caption{The first row is the classes of the source domain and the last row is the classes in the target domain. (a) is an example of consistent and (b)-(c) are the inconsistent taxonomies covered by our CSI. The figure is redrawn from TACS.}
  \label{fig:taxonomy}
\end{figure}

\section{Method}
\subsection{Problem Statement}

Traditional UDA has been conducted on consistent taxonomy where the source and target domains share the same label space, as shown in (a) in Fig. \ref{fig:taxonomy}. However, this assumption is far from being applied to practical problems. To address this limitation, recent studies have been exploring domain adaptation on inconsistent taxonomy. We handle two forms of inconsistent taxonomy settings in the benchmarks of traditional UDA, as shown in (b)-(c) in Fig. \ref{fig:taxonomy}. The first case is \textit{open} taxonomy where classes present in the target domain are not in the source domain. The next case is the \textit{coarse-to-fine} taxonomy. In this case, the classes in the source domain are labeled more finely in the target domain. For instance, a class labeled as \textit{car} in the source domain could be labeled as \textit{car} and \textit{truck} in the target domain. For benchmarks with such inconsistent taxonomies, previous studies have evaluated models only with classes from the source domain. We propose a method that goes beyond the limitation of previous traditional UDA models to predict all classes in the target domain even in mixed inconsistent taxonomy that combines open and coarse-to-fine taxonomy.

\subsection{Base Apporach to the Consistent Taxonomy }

In a consistent taxonomy, traditional UDA with self-training can be divided into two parts: training using images from the source domain and images from the target domain. First, training on the source domain is equivalent to a typical supervised semantic segmentation based on categorical Cross-Entropy (CE). The neural network $f_\theta$ is trained on the source domain image $x^{src}$ consisting of $C$ classes and its ground truth label $y^{src}$. 
\begin{equation}
  \mathcal{L}^{src}_{CE} = -\sum^{C}_{c=1} y^{src}_{c}\log \hat{y}^{src}_{c}
  \label{eq:loss_src}
\end{equation}
Then, the adaptation to the target domain is done through unsupervised learning. The model $f_\theta$ is trained by the pseudo label $p^{trg}$ that is generated by the teacher model $f_\phi$. After each training step $t$, the weights $\phi$ of the teacher model are updated via an Exponential Moving Average (EMA) \cite{tarvainen2017mean} of the weight $\theta$ of the student model $f_\theta$, as shown in Eq. \ref{eq:ema}. The EMA factor $\alpha$ determines the weight of the EMA.
\begin{equation}
  \phi_{t+1} = \alpha(\phi_t -\theta_t) + \theta_t
  \label{eq:ema}
\end{equation}
In general, pseudo labels generated by the teacher model $f_\phi$ at the beginning of training are very noisy and unreliable at the beginning of training. A common strategy to mitigate this is to incorporate a confidence estimate $q$ into the loss function.
\begin{equation}
  \mathcal{L}^{trg}_{CE} = -\sum^{C}_{c=1} q^{trg}_{c}p^{trg}_{c}\log \hat{y}^{trg}_{c}
  \label{eq:loss_trg}
\end{equation}

In this paper, our proposed method works by integrating with self-training UDA methods such as MIC \cite{hoyer2023mic} or DAFormer \cite{hoyer2022daformer}. DAFormer is a method with a Transformer backbone, which is the base method of recent UDA methods \cite{hoyer2023mic, hoyer2022hrda, chen2023pipa} for semantic segmentation. DAFormer uses consistency training with a mixed image following DACS \cite{tranheden2021dacs} as the target image $x^{trg}$. This method also proposed three additional strategies: Rare Class Sampling (RCS), Feature Distance (FD), and Learning Rate Warmup. MIC integrated masked image consistency with HRDA \cite{hoyer2022hrda} which uses multi-crop consistency and is based on DAFormer.

\begin{figure}[tb]
  \centering
  \includegraphics[width=\linewidth]{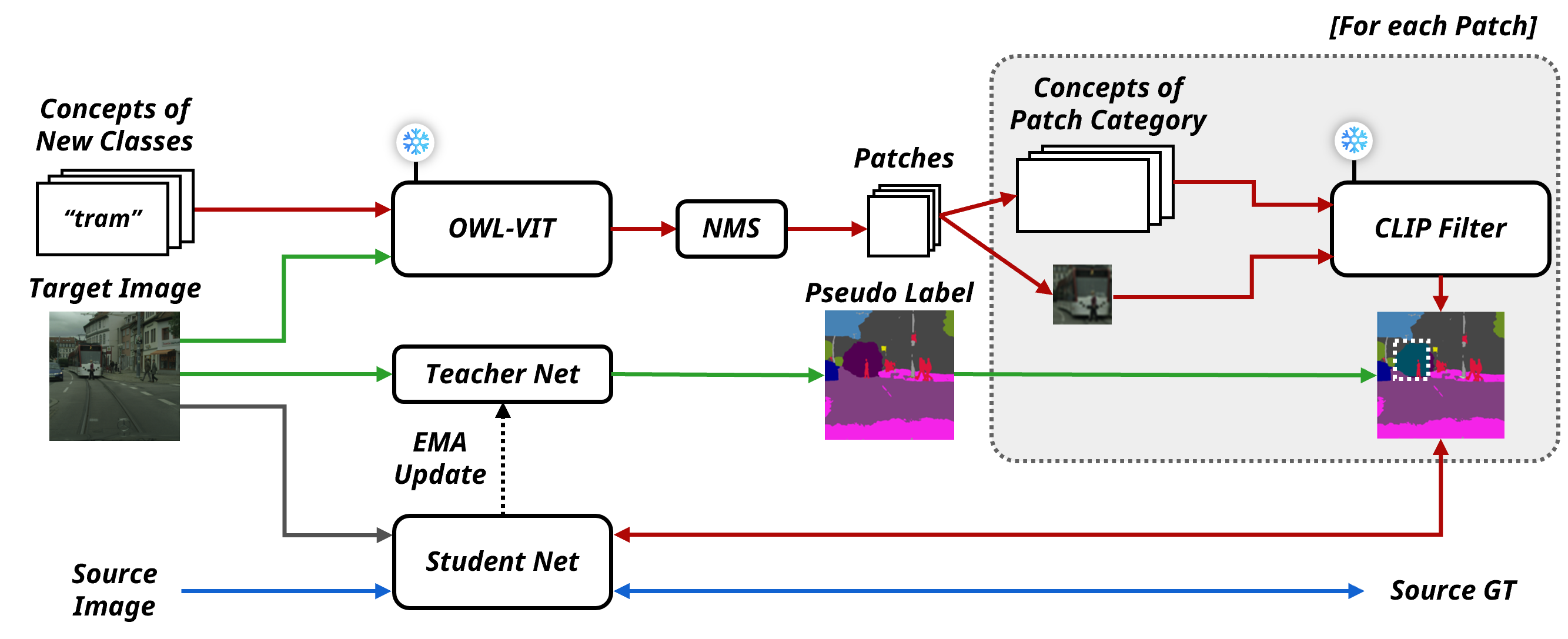}
  \caption{Overview of CSI combined with UDA. The blue line is typically the process of training the model on the source domain. The gray line is the process of adapting the model to the target domain. The red line is the process of adapting the model to the target domain with our proposed method.}
  \label{fig:overview}
\end{figure}

\subsection{Our Approach to the Inconsistent Taxonomy}
\subsubsection{Motivation}

In the traditional UDA with inconsistent taxonomy, the model can't learn classes that are only in the target domain because neither the source domain nor the target domain is labeled for those classes. We found consistent characteristics in the predictions of models trained with traditional UDA on inconsistent taxonomy. New classes in the target domain were predicted as classes with a similar appearance or texture in the source domain. This means that those classes in the target domain are overfitted with specific classes in the source domain. This has even been observed in previous studies in the traditional UDA with consistent taxonomy. For example, there was a problem with the DAFormer network predicting \textit{train} class as \textit{bus} class after a few hundred training steps in DAFormer. The authors of DAFormer assumed that the model was overfitted to the source domain and added Feature Distance (FD) loss using a model pre-trained with ImageNet to mitigate this problem. We try to use this phenomenon to solve the problem of inconsistent taxonomies.

We consider the process of semantic segmentation of an image by a model to be divided into two types of reasoning processes: segment reasoning and semantic reasoning. First, segment reasoning is the inference process for the model to segment the image. Semantic reasoning is the inference to assign each segment to the predicted class. As learning progresses with traditional UDA methods, the accuracy of spatial segment reasoning for the target image improves, but since the source-target taxonomy is different, the pseudo labels generated by UDA are limited to the source taxonomy. Without correct labels for the target data, it is challenging to perform accurate semantic reasoning. If incorrect semantic reasoning can be properly corrected, the UDA model will be able to rectify classes it confuses. 

\subsubsection{Procedure Overview}

In this paper, we assume that the taxonomies (individual class names) of source and target are known. By using VLM models to perform object detection on classes that only exist in the target domain, we can contrast the detected areas with the pseudo labels obtained from UDA segment results, creating an opportunity to correct the labels. For example, if only the \textit{bus} class exists in the source domain and not the \textit{train} class, the \textit{train} of the target domain will be segmented as \textit{bus} in the pseudo labels. However, by using VLMs to perform zero-shot object detection for the \textit{train} class in the target image, we can increase the opportunity for semantic reasoning for the \textit{train} class. Nevertheless, using zero-shot object detection alone is not sufficient, so we can improve zero-shot relabeling performance by proceeding in stages as follows. Fig. \ref{fig:overview} provides an overview of this process. 

\subsubsection{Build a Map to Relabel}
We can create a mapping structure, Map, to resolve the inconsistency in taxonomy. In the coarse-to-fine taxonomy scenario, it's possible to identify in advance which classes in the source domain are split into sub-classes in the target domain. This enables us to store the relationship information in a Map.From-To entry. For example, if the class \textit{vegetation} from the source domain is divided into \textit{vegetation} and \textit{terrain} in the target domain, we can assign \textit{vegetation} to `Map.From' and \textit{terrain} to `Map.To'. However, under the open taxonomy scenario, it is difficult to know in advance the From-To relationships for classes that newly appear in the target domain. To overcome this, we propose a method for automatically constructing mapping relationships during the training process.

\subsubsection{Extract Patches}

We leverage the sophisticated semantic reasoning capabilities of Vision-Language Models (VLMs) to extract semantic information for classes that are present in the target domain but absent from the source domain. A zero-shot object detection model trained on extensive open-vocabulary data can address the class taxonomy inconsistency between source and target domains by detecting objects of arbitrary classes. In this study, we employ the OWL-ViT\cite{minderer2022simple} model for zero-shot object detection on target images, specifically targeting classes that are exclusive to the target domain. To ensure accuracy, objects detected with confidence levels below a predetermined threshold are discarded. Furthermore, we apply Non-Maximum Suppression (NMS) to remove overlapping detected objects, thereby enhancing the precision of our object detection process. The culmination of this process involves cropping the detected object's bounding box from both the target image and the pseudo label generated via UDA, subsequently saving it as a patch. However, our findings reveal limitations in the effectiveness of OWL-ViT for zero-shot object detection, as demonstrated in Tab. \ref{tab:ab_ca}. These limitations arise primarily from the model's inadequate training on target domain data and the difficulty of distinguishing among numerous similar objects in the dataset. For instance, the model frequently misclassifies vans, which belong to the \textit{car} class, as members of the \textit{truck} class when attempting to identify the \textit{truck} class in target domain images.

\subsubsection{Filter Patches}

To address the issue mentioned, we can utilize another VLM model, CLIP, to filter out inaccurately detected patches. By employing CLIP, we perform zero-shot classification on the cropped target image contained in each patch. This process allows us to retain only the patches corresponding to the target domain classes we are interested in. For instance, if OWL-ViT tries to detect a \textit{truck} class, but obtains a \textit{van} object that actually belongs to the \textit{car} class, we can use CLIP to verify that it is not a \textit{truck} but a \textit{van}, thus excluding that patch. The score value for each class output is calculated as a probability using the softmax function, and the class with the highest probability is the predicted class. Only if the predicted class is the same as the class detected by the zero-shot object detection model and is above the threshold, the patch is considered correctly detected. We can get more accurate patches through this process.

\begin{equation}
  y^{cls} = \arg \max(softmax(f_{clip}(x^{img}, x^{txt}))
  \label{eq:clip}
\end{equation}

Moreover, by examining the cropped pseudo label included in each patch obtained by OWL-ViT, we can assess the necessity for relabeling. If the cropped pseudo label does not contain any class of all the Map.From entries, it indicates there are no pixels requiring relabeling, and thus the patch can be excluded. Performing this step before the CLIP-based zero-shot classification process can efficiently reduce unnecessary CLIP inference computations.

\subsubsection{Relabel and Paste Patches}

Once the patches are accurately extracted and filtered using VLM models, the relabeling and pasting process can be simplified as follows. Based on the information on the Map.From-To entries, the class of Map.From contained in each patch's cropped pseudo label is replaced with the Map.To class information. This modified version of the cropped pseudo label is then pasted back onto the original (uncropped) pseudo label, and the UDA student network is trained with this updated information.

\subsubsection{Auto-Configure a Map}
However, for new classes in an open taxonomy, it is difficult to generate a Map.From-To entry before starting training. To solve this, we design a process to find relations to generate a Map.From-To entry during training. First, we train the model with the traditional UDA method up to 8K training steps to generate pseudo labels that are not perfect but are somewhat reliable. Then, from 8K to 12K training steps, we detect the Map.To class and extract patches (image and pseudo label) for the detected region. For each pseudo label patch, we find the class that exceeds the threshold and occupies the most area at the pixel level. These classes are candidates to become Map.From entry. After 12K training steps, the most counted class that has the same class category as Map.To class is selected as a Map.From entry. Finally, we generate a Map.From-to entries including the selected a Map.From entry. From 12K training steps, relabeling is performed using the generated Map.From-To entries.

\subsubsection{Convert Class to Concept}
Inspired by SemiVL\cite{hoyer2023semivl}, we construct a set of concepts for each class, which include additional subcategories or descriptions extracted from the class definitions. Using these concepts allows us to extract and filter patches more accurately. For example, the concept of \textit{Car}, \textit{SUV}, \textit{jeep}, and \textit{Van} can belong to the \textit{Car} class. Therefore, the text inputs $x^{txt}$ of OWL-ViT and CLIP are expanded to a set of concepts. The concepts used in the experiment are shown in the supplement.

\begin{algorithm}
\caption{CSI algorithm}
\label{alg:cap}
\algnewcommand\algorithmicto{\textbf{to}}
\renewcommand{\algorithmicrequire}{\textbf{Input:}}
\renewcommand{\algorithmicensure}{\textbf{Output:}}
\begin{algorithmic}
\Require Current step: $i$, Target image: $x$, Pseudo-label: $\hat{y}$, Relabeling Map: $M$
\Ensure Relabeled Pseudo-label: $\hat{y}$

\If {$i$ in relabeling step}
    \State $P \gets$ \Call{Extract}{$x, \hat{y}, M$} \Comment{ Zero-shot object detection by OWL-ViT}    
    \State $P \gets$ \Call{Filter}{$P$} \Comment{Zero-shot classification by CLIP}    
    \ForAll {$p \in P$}        
        \If {$p$ contains any Map.From classes}
            \State $p \gets $ Relabel($p$)
            \State $\hat{y} \gets $ Paste($p, \hat{y}$)
    \EndIf
\EndFor
\EndIf
\State \Return $\hat{y}$

\end{algorithmic}
\end{algorithm}

\section{Experiments}
\subsection{Implementation Details}
\subsubsection{Datasets}
Like previous UDA methods, we use Synthia \cite{ros2016synthia} and GTA \cite{richter2016playing} as source domains and Cityscapes \cite{cordts2016cityscapes} as the target domain. In addition, we use ACDC \cite{sakaridis2021acdc} as a more challenging target domain. Cityscapes is a dataset of European street scenes collected from a car in clear weather. It consists of 2,975 images for training and 500 images for validation with a resolution of 2048×1024. We use a format where the labels are organized into 19 classes. Synthia was collected in a simulated environment and consists of 9,400 images with a resolution of 1280×760. It is labeled into 16 classes, which are a subset of Cityscapes. GTA was collected from Grand Theft Auto V, one of the most popular open-world games. It consists of 24,966 images. Finally, we use ACDC, which consists of 1,600 images for training and 406 images for validation with a resolution of 1920×1080. GTA and ACDC have the same class taxonomy as Cityscapes.

\subsubsection{Network Architecture}
To evaluate our proposed method, we utilize MIC \cite{hoyer2023mic} combined with HRDA \cite{hoyer2022hrda} and DAFormer \cite{hoyer2022daformer} as our base method. DAFormer consists of a SegFormer \cite{xie2021segformer} encoder and a context-aware fusion decoder. We use OWL-ViT \cite{minderer2022simple} based on ViT-Base with patch size 32 for zero-shot object detection. CLIP \cite{radford2021learning} is also based on the same model and patch size. When applying the proposed method in Tab. \ref{tab:ab_diff_uda} to different UDA methods, DeepLabV2 uses ResNet101 as a backbone.

\subsubsection{Training}
By default, we follow the training process applied to MIC with HRDA. The MIC with the proposed method is trained for 40K iterations with batch size 2. We use linear learning rate warmup for the first 1.5K iteration and adopt AdamW as the optimizer with a learning rate of $6\times10^{-5}$ for the encoder and $6\times10^{-4}$ for the decoder. We set the mask ratio to 0.7 and the mask block to 64. Collecting info to generate a relabeling map starts 4K steps before relabeling and ends before relabeling. Relabeling starts at 12K iterations which is the point where the performs best in Fig. \ref{fig:relabel_start}. The detection thresholds for \textit{terrain}, \textit{truck}, and \textit{train} in OWL-ViT are 0.01, 0.1, and 0.1, respectively. The classification thresholds for \textit{terrain}, \textit{truck}, and \textit{train} in CLIP are 0.1, 0.5, and 0.5, respectively.

\begin{table}[tb]
  \caption{Quantitative comparison with previous UDA methods on Synthia-to-Cityscapes. The results are averaged over 3 random seeds. We calculated mIoU${_{19}}$ by setting 0 for classes that are not predicted by traditional UDA. The classes with gray (\textit{terrain}, \textit{truck}, and \textit{train}) are not in the source domain.
  }
  \label{tab:syn2city_sota}
  \centering\resizebox{\textwidth}{!}{
  \begin{tabular}{@{}lccccccccc>{\columncolor[gray]{0.8}}ccccc>{\columncolor[gray]{0.8}}cc>{\columncolor[gray]{0.8}}ccc|cc@{}}
    \toprule
    Method & Road & S.walk & Build. & Wall & Fence & Pole & Tr.Light & Sign & Veget. & Terrain & Sky & Person & Rider & Car & Truck & Bus & Train & M.bike & Bike & mIoU$_{16}$ & mIoU$_{19}$ \\
    \midrule
    \multicolumn{22}{c}{Sim-to-Real: Synthia to Cityscapes (Val.)} \\
    \midrule
    CBST & 68.0 & 29.9 & 76.3 & 10.8 & 1.4 & 33.9 & 22.8 & 29.5 & 77.6 & - & 78.3 & 60.6 & 28.3 & 81.6 & - & 23.5 & - & 18.8 & 39.8 & 42.6 & 35.9 \\
    DACS & 80.6 & 25.1 & 81.9 & 21.5 & 2.9 & 37.2 & 22.7 & 24.0 & 83.7 & - & 90.8 & 67.6 & 38.3 & 82.9 & - & 38.9 & - & 28.5 & 47.6 & 48.3 & 40.7 \\
    CorDA & \textbf{93.3} & \textbf{61.6} & 85.3 & 19.6 & 5.1 & 37.8 & 36.6 & 42.8 & 84.9 & - & 90.4 & 69.7 & 41.8 & 85.6 & - & 38.4 & - & 32.6 & 53.9 & 55.0 & 46.3 \\
    ProDA & 87.8 & 45.7 & 84.6 & 37.1 & 0.6 & 44.0 & 54.6 & 37.0 & 88.1 & - & 84.4 & 74.2 & 24.3 & 88.2 & - & 51.1 & - & 40.5 & 45.6 & 55.5 & 46.7 \\
    DAFormer & 84.5 & 40.7 & 88.4 & 41.5 & 6.5 & 50.0 & 55.0 & 54.6 & 86.0 & - & 89.8 & 73.2 & 48.2 & 87.2 & - & 53.2 & - & 53.9 & 61.7 & 60.9 & 51.3 \\
    DAFormer\texttt{+}CSI & 80.0 & 38.2 & 88.6 & 40.7 & 6.7 & 50.0 & 55.7 & 52.0 & 84.5 & 23.6 & 91.8 & 72.7 & 45.7 & 90.5 & 70.1 & 75.9 & 64.7 & 51.3 & 57.8 & 61.4 & 60.0 \\
    \midrule
    HRDA & 85.2 & 47.7 & 88.8 & 49.5 & 4.8 & 57.2 & 65.7 & 60.9 & 85.3 & - & 92.9 & 79.4 & 52.8 & 89.0 & - & 64.7 & - & 63.9 & 64.9 & 65.8 & 55.4 \\
    MIC & 86.6 & 50.5 & 89.3 & 47.9 & 7.8 & \textbf{59.4} & 66.7 & \textbf{63.4} & 87.1 & - & \textbf{94.6} & \textbf{81.0} & \textbf{58.9} & 90.1 & - & 61.9 & - & \textbf{67.1} & 64.3 & 67.3 & 56.7 \\
    PiPa & 88.6 & 50.1 & \textbf{90.0} & \textbf{53.8} & 7.7 & 58.1 & \textbf{67.2} & 63.1 & \textbf{88.5} & - & 94.5 & 79.7 & 57.6 & 90.8 & - & 70.2 & - & 65.1 & \textbf{66.9} & 68.2 & 57.5 \\
    \midrule
    MIC\texttt{+}CSI & 88.1 & 52.9 & 89.3 & 46.0 & \textbf{8.3} & 59.2 & 66.1 & 60.9 & 84.1 & \textbf{28.6} & \textbf{94.6} & \textbf{81.0} & 56.6 & \textbf{94.1} & \textbf{76.7} & \textbf{87.0} & \textbf{74.5} & 66.5 & 66.0 &  \textbf{68.8} & \textbf{67.4} \\
    \bottomrule
  \end{tabular}}
\end{table}

\begin{figure}[tb]
  \centering
  \includegraphics[width=\textwidth]{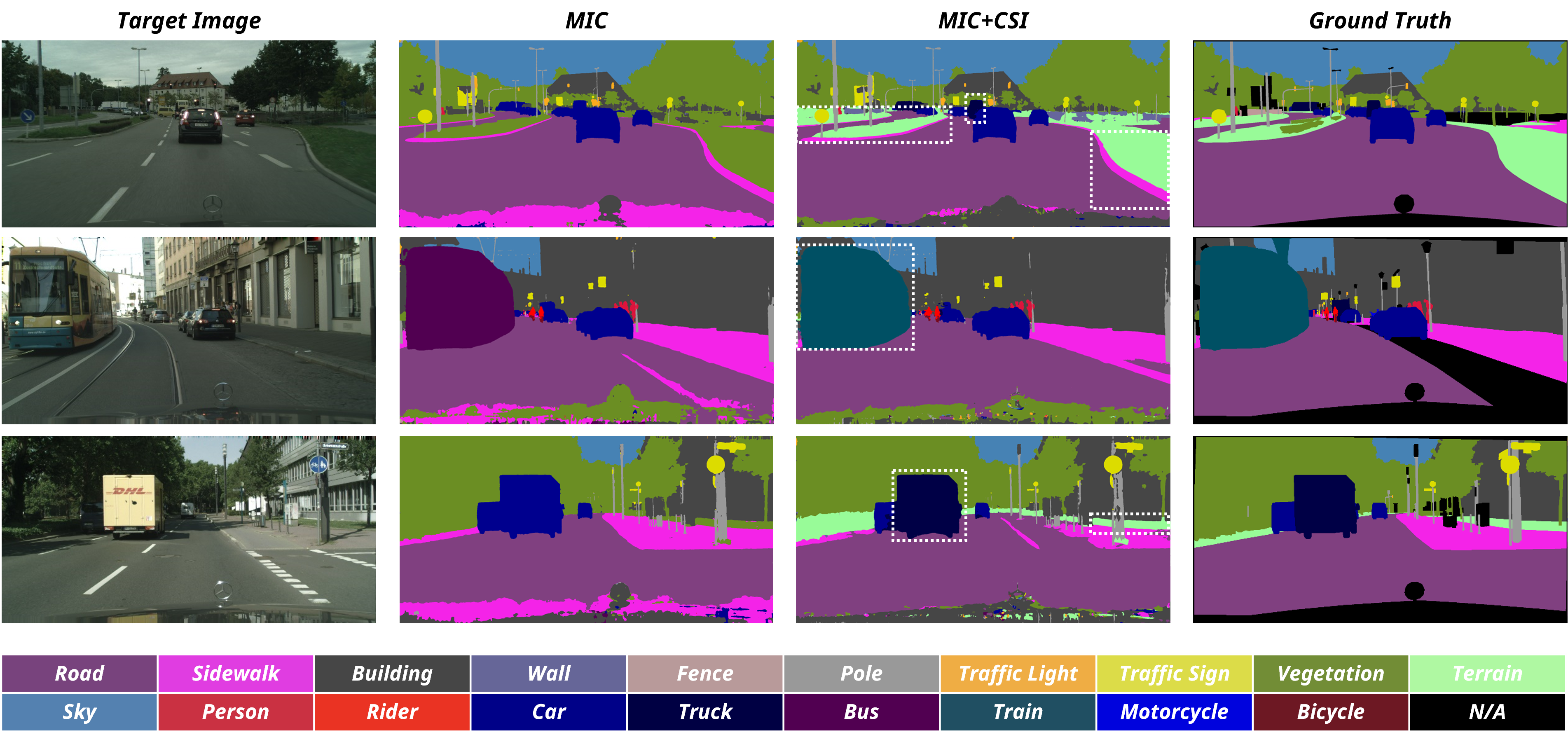}
  \caption{Qualitative comparison of CSI with previous methods on Synthia-to-Cityscapes. CSI shows better segmentation performance for classes not in the source domain.}
  \label{fig:qc_sota}
\end{figure}

\subsection{Comparison with the State of the Art}
We compare the proposed method to previous traditional UDA methods on Synthia-to-Cityscapes.
Previously, Synthia and Cityscapes were considered to share the same label space. Therefore, previous studies evaluate their models against Synthia with 16 classes. We depart from this limitation by applying our proposed method to the 3 classes that are unique to Cityscapes and evaluate the model for 19 classes. The classes that are only present in Cityscapes are \textit{terrain}, \textit{truck}, and \textit{train}. The \textit{vegetation} in Synthia is labeled as \textit{vegetation} and \textit{terrain} classes in Cityscapes, and the \textit{car} class in Synthia is divided into \textit{car} and \textit{truck} in Cityscapes. Therefore, the domain adaptation for those classes corresponds to the coarse-to-fine taxonomy setting. We use the predefined map for both classes for relabeling. On the other hand, \textit{train} in cityscapes is an object that does not exist in Synthia. In this case, it is in the open taxonomy setting. Since we cannot get a predefined map for the class in the open taxonomy, we use the map generated by the proposed method to relabel it.

In Tab. \ref{tab:syn2city_sota}, we evaluate our proposed method not only with 19 classes but also with 16 classes. The proposed method outperforms the previous best state-of-the-art method by \texttt{+}0.6 mIoU when evaluated with 16 classes. When evaluated with 19 classes, it outperforms the previous traditional UDA by \texttt{+}9.9 mIoU. Our proposed method is only applied to 3 classes that are not in the source domain, but it can also improve the performance of consistent classes in the target domain. In particular, the performance improvement is high for classes related to relabeling. This can be seen visually in Fig. \ref{fig:qc_sota}.

\begin{table}[tb]
  \setlength{\tabcolsep}{4pt}
  \caption{Ablation of the different UDA methods on Synthia-to-Cityscapes.}
  \label{tab:ab_diff_uda}
  \centering{
  \begin{tabular}{cll|cccc}
    \toprule
    & \multirow{2}{*}{Method} & \multirow{2}{*}{Base} & \multicolumn{2}{c}{mIoU$_{16}$} & \multicolumn{2}{c}{mIoU$_{19}$} \\
    &  &  & w/o CSI & w/ CSI & w/o CSI & w/ CSI \\
    \midrule
    1 & DACS & DeepLabV2 & 48.3 & 48.9 & 40.7 & 45.5 \\
    2 & DAFormer & DeepLabV2 & 51.7 & 51.8 & 43.6 & 49.6 \\
    3 & DAFormer & DAFormer & 60.9 & 61.4 & 51.3 & 60.0 \\
    4 & HRDA & DAFormer & 65.8 & 66.7 & 55.4 & 64.9 \\
    5 & MIC & HRDA & 67.3 & 68.8 & 56.7 & 67.4 \\
  \bottomrule
  \end{tabular}}
\end{table}

\subsection{Applying to Other UDA Methods}

Our proposed method can be easily applied to other UDA methods. In Tab. \ref{tab:ab_diff_uda}, we apply our proposed method to other UDA methods on Synthia-to-Cityscapes and compare the results on 16 classes and 19 classes. Our proposed method consistently improves the performance for 16 classes, while also improving the performance for 19 classes. In particular, we can see that our proposed method performs well not only on transformer-based models but also on CNN-based models.

\begin{figure}[tb]
  \centering
  \includegraphics[width=1.0\linewidth]{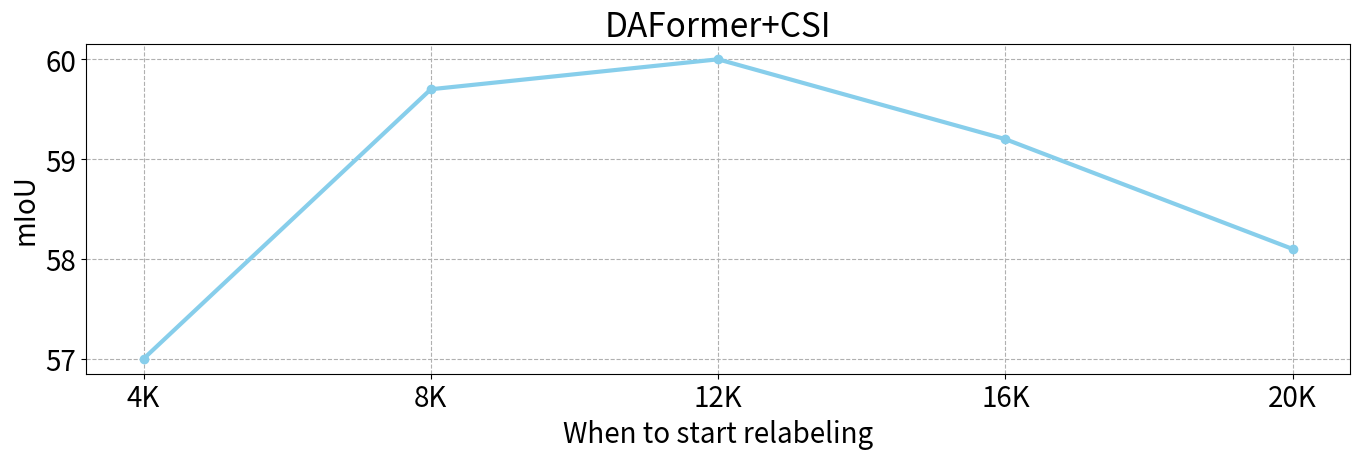}
  \caption{Performance for when relabeling started on Synthia-to-Cityscapes. The performance improvement decreases as the relabeling time is delayed.}
  \label{fig:relabel_start}
\end{figure}

\subsection{Influence of When Relabeling Starts }
Our proposed method is applied during training with the traditional UDA method that the model is based on. We conduct an ablation study on the relabeling time to determine the exact time to start relabeling. Fig. \ref{fig:relabel_start}, if we start relabeling too late, the model will have less time to learn about the relabeled classes, resulting in less performance improvement. Also, if the model starts relabeling too early, it has difficulty finding which class the model has overfitted for the new class.

\begin{table}[tb]
  \setlength{\tabcolsep}{4pt}
  \caption{Ablation study of the components on Synthia-to-Cityscapes.}
  \label{tab:ab_ca}
  \centering{
  \begin{tabular}{ccc|ccc|cccc}
    \toprule
    & OWL-ViT & CLIP & Terrain & Truck & Train & \multicolumn{2}{c}{mIoU$_{16}$} & \multicolumn{2}{c}{mIoU$_{19}$}\\
    \midrule
    1 & - & - & - & - & - & 67.3 & \raisebox{1ex}{\rotatebox{270}{$\Rsh$}} & 56.7 & \raisebox{1ex}{\rotatebox{270}{$\Rsh$}} \\
    2 & \checkmark & - & 12.6 & 15.0 & 68.0 & 64.8 & \texttt{-}2.5 & 60.6 & \texttt{+}3.9 \\
    3 & \checkmark & \checkmark & 28.6 & 76.7 & 74.5 & 68.8 & \texttt{+}1.5 & 67.4 & \texttt{+}10.7 \\
  \bottomrule
  \end{tabular}}
\end{table}

\subsection{Component Ablations}

We ablated the components of our proposed method in Tab. \ref{tab:ab_ca}. Our model consists of two main components: OWL-ViT for zero-shot object detection and CLIP for validating the detected patches. By default, even if we only use OWL-ViT, the model can relabel the detected patches. However, in this case, there is a performance degradation due to incorrectly detected patches. In particular, the \textit{truck} class shows very poor performance when CLIP is not used. As a result, the performance for the 16 classes also drops by -2.5 mIoU. Therefore if the performance of the zero-shot object detection model improves and it can detect the classes more accurately, we can expect more performance improvement.

\begin{table}[tb]
  \caption{Quantitative comparison with previous UDA methods on GTA-to-Cityscapes.}
  \label{tab:ab_gta2cs}
  \centering\resizebox{\textwidth}{!}{
  \begin{threeparttable}
  \begin{tabular}{@{}lccccccccc>{\columncolor[gray]{0.8}}ccccc>{\columncolor[gray]{0.8}}cc>{\columncolor[gray]{0.8}}ccc|cc@{}}
    \toprule
    Method & Road & S.walk & Build. & Wall & Fence & Pole & Tr.Light & Sign & Veget. & Terrain & Sky & Person & Rider & Car & Truck & Bus & Train & M.bike & Bike & mIoU$_{16}$ & mIoU$_{19}$ \\
    \midrule
    \multicolumn{22}{c}{Game-to-Real: GTA to Cityscapes (Val.)} \\
    \midrule
    DAFormer$^*$ & 95.5 & 68.5 & 89.5 & 53.2 & 47.3 & 50.2 & 55.4 & 60.1 & 87.1 & - & 91.0 & 72.1 & 44.9 & 89.0 & - & 63.8 & - & 53.7 & 62.1 & 67.7 & 57.0 \\
    DAFormer$^*$\texttt{+}CSI & 93.2 & 60.7 & 89.4 & 53.4 & 48.0 & 50.6 & 55.4 & 60.6 & 88.5 & 41.5 & 90.6 & 71.7 & 44.0 & 90.6 & 42.1 & 73.5 & 63.0 & 55.3 & 61.2 & 67.9 & 64.9 \\
    DAFormer & 95.7 & 70.2 & 89.4 & 53.5 & 48.1 & 49.6 & 55.8 & 59.4 & 89.9 & 47.9 & 92.5 & 72.2 & 44.7 & 92.3 & 74.5 & 78.2 & 65.1 & 55.9 & 61.8  & 69.3 & 68.3 \\
    \midrule
    MIC$^*$ & 97.4 & 80.4 & 91.8 & 61.2 & 55.2 & 60.0 & 65.5 & 73.0 & 89.3 & - & 93.1 & 79.5 & 52.8 & 90.6 & - & 68.2 & - & 64.7 & 67.0 & 74.4 & 62.6 \\
    MIC$^*$\texttt{+}CSI & 97.3 & 80.0 & 91.8 & 62.1 & 54.4 & 58.9 & 65.2 & 72.4 & 90.0 & 39.2 & 92.8 & 79.5 & 54.0 & 93.0 & 68.0 & 84.1 & 74.8 & 64.5 & 67.4 & 75.5 & 73.1 \\
    MIC & 97.4 & 80.1 & 91.7 & 61.2 & 56.9 & 59.7 & 66.0 & 71.3 & 91.7 & 51.4 & 94.3 & 79.8 & 56.1 & 94.6 & 85.4 & 90.3 & 80.4 & 64.5 & 68.5 & 76.5 & 75.9 \\
    \bottomrule
  \end{tabular}
\begin{tablenotes}       
    \item[*] Model trained without \textit{terrain}, \textit{truck}, and \textit{train} on the source domain.
\end{tablenotes}
\end{threeparttable}}
\end{table}

\begin{table}[tb]
  \caption{Quantitative comparison with previous UDA methods on Synthia-to-ACDC.}
  \label{tab:ab_syn2acdc}
  \centering\resizebox{\textwidth}{!}{
  \begin{tabular}{@{}lccccccccc>{\columncolor[gray]{0.8}}ccccc>{\columncolor[gray]{0.8}}cc>{\columncolor[gray]{0.8}}ccc|cc@{}}
    \toprule
    Method & Road & S.walk & Build. & Wall & Fence & Pole & Tr.Light & Sign & Veget. & Terrain & Sky & Person & Rider & Car & Truck & Bus & Train & M.bike & Bike & mIoU$_{16}$ & mIoU$_{19}$ \\
    \midrule
    \multicolumn{22}{c}{Sim-to-Adverse: Synthia to ACDC (Val.)} \\
    \midrule
    DAFormer & 58.0 & 16.6 & 62.3 & 22.0 & 4.9 & 41.4 & 28.6 & 34.3& 64.0 & - & 78.8 & 46.5 & 15.3 & 68.7& - & 29.4 & - & 24.0 & 3.9 & 37.4 & 31.5 \\
    HRDA & 70.4 & 5.9 & 68.8 & 22.5 & 3.3 & 57.2 & 32.5 & 48.2 & 72.2 & - & 79.4 & 57.2 & 25.2 & 69.2 & - & 40.2 & - & 22.9 & 5.9 &   42.6 & 35.8 \\
    MIC & 65.2 & 7.2 & 73.6 & 37.2 & 4.4 & 58.3 & 43.5 & 45.2 & 68.9 & - & 73.2 & 62.8 & 36.7 & 69.3 & - & 39.9 & - & 23.5 & 6.8 & 44.7 & 37.7 \\
    MIC\texttt{+}CSI & 64.8 & 10.1 & 70.4 & 43.5 & 4.8 & 60.1 & 25.3 & 49.2 & 69.1 & 20.2 & 70.6 & 64.2 & 39.8 & 79.8 & 74.9 & 50.1 & 58.6 & 26.4 & 10.9 & 46.2 & 47.0 \\    
    \bottomrule
  \end{tabular}}
\end{table}

\subsection{Applying to different Domain Configurations}

We evaluate our proposed method for more diverse taxonomy settings and domains. First, the domain adaptation from GTA to Cityscapes is a very common benchmark in UDA. We convert the labels of GTA to evaluate our method on that benchmark. To match the same taxonomy as Synthia as much as possible, we changed the \textit{terrain}  to \textit{vegetation}, \textit{truck} to \textit{car}, and \textit{train} to \textit{unlabeled} of GTA. The taxonomy that differs between GTA-to-Cityscapes and Synthia-to-Cityscapes is that the \textit{train} in GTA-to-Cityscapes is unlabeled and the \textit{train} in Synthia-to-Cityscapes is unseen. As shown in Tab. \ref{tab:ab_gta2cs}, \textit{train} achieves 93\% of the performance of \textit{train} on fully labeled GTA even though it is unlabeled on converted GTA. However, even using a source domain similar to the target domain did not necessarily result in higher performance on the new classes.

As mentioned in the previous section, our model performs better the more accurately it detects objects. Therefore, adjusting the threshold of the model to the target domain shows better performance. We evaluated our method by keeping the original settings and changing the target domain to a more challenging environment, ACDC, to see if our method still performs well without necessarily optimizing the values for the target domain. As shown in Tab. \ref{tab:ab_syn2acdc}, our proposed method shows \texttt{+}1.5 mIoU higher performance compared to baseline existing models evaluated with 16 classes. In particular, the performance of \textit{car} is \texttt{+}10.5 mIoU better than the previous method. This shows that the segment reasoning ability of \textit{car} helps to segment \textit{truck} well.

\section{Conclusion}

In this paper, we propose a method to apply the powerful semantic capabilities of VLM-based zero-shot models to traditional UDA methods through relabeling to perform UDA in a mixed setting of coarse-to-fine and open taxonomies. It can be used in combination with several traditional UDA methods and showed consistent performance improvements. For Synthia-to-Cityscapes, CSI combined with MIC showed a performance increase of \texttt{+}9.9 mIoU on inconsistent taxonomy over Synthia-to-Cityscapes. At the same time, there is a small performance increase in the original classes. We also demonstrate notable performance on classes that are only in the target domain on benchmark datasets of various configurations. Concerns and limitations of the CSI are discussed in the supplement.

\subsubsection{Acknowledges}

This work was partly supported by the National Research Foundation of Korea(NRF) grant funded by the Korea government (MSIT) (No. RS-2024-00348376) and the Institute of Information \& Communications Technology Planning \& Evaluation (IITP) grant funded by the Korea government (MSIT) (No. RS-2024-00438686, Development of software reliability improvement technology through identification of abnormal open sources and automatic application of DevSecOps), (No. 2022-0-01045, Self-directed Multi-Modal Intelligence for solving unknown open domain problems), (No. RS-2022-II220688, AI Platform to Fully Adapt and Reflect Privacy-Policy Changes), and (No. 2019-0-00421, Artificial Intelligence Graduate School Program).

%
%
\bibliographystyle{splncs04}
\bibliography{egbib}

\newpage
\appendix

\section*{Supplementary Material}
\setcounter{table}{0}
\renewcommand{\thetable}{S\arabic{table}}

\setcounter{figure}{0}
\renewcommand{\thefigure}{S\arabic{figure}}

\section{Overview}

In the supplementary material, we provide trivial experimental environment details (Sec. \ref{sec:experimental}), the hyper-parameters used in experiments (Sec. \ref{sec:parameters}), further analysis (Sec. \ref{sec:rep_analysis}-\ref{sec:qual_analysis}) and discussion (Sec. \ref{sec:discussion}).

\section{Experimental Environment}
\label{sec:experimental}

CSI works by integrating with DAFormer \cite{hoyer2022daformer} or MIC \cite{hoyer2023mic}. The experimental environments in MIC and DAFormer are a bit outdated. For example, MIC performs best with Pytorch version 1.7.1 and CUDA version 11.0 and uses mmsegmentation version 0.16.0. To utilize up-to-date models and for future research, we reproduced the MIC code with the latest Pytorch (v2.1.2) and mmsegmentation frameworks (v1.2.2). Tab. \ref{tab:ab_repr} shows the performance of DAFormer and MIC trained with our reproduced code on Synthia-to-Cityscapes. Our experiments are conducted on an RTX 4090 or RTX A6000 GPU. The source code is available at \url{https://github.com/jkee58/CSI}.

\begin{table}[!ht]
  \caption{Quantitative comparison of DAFormer and MIC with and without reproduction on Synthia-to-Cityscapes.}
  \label{tab:ab_repr}
  \centering\resizebox{\textwidth}{!}{
  \begin{tabular}{@{}lccccccccc>{\columncolor[gray]{0.8}}ccccc>{\columncolor[gray]{0.8}}cc>{\columncolor[gray]{0.8}}ccc|cc@{}}
    \toprule
    Method & Road & S.walk & Build. & Wall & Fence & Pole & Tr.Light & Sign & Veget. & Terrain & Sky & Person & Rider & Car & Truck & Bus & Train & M.bike & Bike & mIoU$_{16}$ & mIoU$_{19}$ \\
    \midrule
    \multicolumn{22}{c}{Sim-to-Real: Synthia to Cityscapes (Val.)} \\
    \midrule
    DAFormer (Paper) & 84.5 & 40.7 & 88.4 & 41.5 & 6.5 & 50.0 & 55.0 & 54.6 & 86.0 & - & 89.8 & 73.2 & 48.2 & 87.2 & - & 53.2 & - & 53.9 & 61.7 & 60.9 & 51.3 \\
    DAFormer (Reproduced) & 80.0 & 37.4 & 87.9 & 41.0 & 7.4 & 50.0 & 48.8 & 52.0 & 85.7 & - & 88.9 & 72.0 & 47.6 & 86.6 & - & 61.6 & - & 55.3 & 56.6 & 59.9 & 50.5 \\
    DAFormer\texttt{+}CSI & 80.0 & 38.2 & 88.6 & 40.7 & 6.7 & 50.0 & 55.7 & 52.0 & 84.5 & 23.6 & 91.8 & 72.7 & 45.7 & 90.5 & 70.1 & 75.9 & 64.7 & 51.3 & 57.8 & 61.4 & 60.0 \\
    \midrule
    MIC (Paper) & 86.6 & 50.5 & \textbf{89.3} & \textbf{47.9} & 7.8 & 59.4 & 66.7 & 63.4 & 87.1 & - & \textbf{94.6} & 81.0 & 58.9 & 90.1 & - & 61.9 & - & \textbf{67.1} & 64.3 & 67.3 & 56.7 \\
    MIC (Reproduced) & 86.6 & 51.7 & 88.2 & 47.8 & 6.2 & \textbf{60.6} & \textbf{67.8} & \textbf{64.1} & \textbf{87.7} & - & 91.0 & \textbf{81.3} & \textbf{60.5} & 90.5 & - & 67.3 & - & 67.0 & 64.8 & 67.7 & 57.0 \\
    MIC\texttt{+}CSI & \textbf{88.1} & \textbf{52.9} & \textbf{89.3} & 46.0 & \textbf{8.3} & 59.2 & 66.1 & 60.9 & 84.1 & \textbf{28.6} & \textbf{94.6} & 81.0 & 56.6 & \textbf{94.1} & \textbf{76.7} & \textbf{87.0} & \textbf{74.5} & 66.5 & \textbf{66.0} &  \textbf{68.8} & \textbf{67.4} \\
    \bottomrule
  \end{tabular}}
\end{table}

\section{Hyper-parameters}
\label{sec:parameters}

CSI utilizes OWL-ViT \cite{minderer2022simple} and CLIP \cite{radford2021learning} to detect new classes that are only in the target domain. In this case, it is necessary to set hyper-parameters in the process of detecting new classes. We set the detection score threshold to the default value of 0.1 for \textit{truck} and \textit{train} as shown in Tab. \ref{tab:parameters}. \textit{Terrain} is a non-object class, so a much lower threshold is required to detect it. The IoU threshold used in NMS is set to 0.3. For classification with CLIP, we used the common threshold of 0.5 for \textit{truck} and \textit{train}. \textit{Terrain} used a much lower threshold for detection. Because \textit{terrain} utilizes a very low threshold, the final extracted patches are often inaccurate. However, this is somewhat mitigated by the fact that we use the map to relabel it.

\begin{table}[!ht]
\centering
\setlength{\tabcolsep}{4pt}
\caption{Hyper-parameters for zero-shot detection and classification.}
\label{tab:parameters}
\begin{tabular}{@{}lll@{}} 
\toprule
Class & Det. thresh. & Cls. thresh.\\
\midrule
Terrain & 0.01 & 0.1 \\
Truck   & 0.1  & 0.5 \\
Train   & 0.1  & 0.5 \\
\bottomrule
\end{tabular}
\end{table}

Since we use open-vocabulary models for detection and classification, the prompt is crucial for patch extraction. Typing the concept of the class is more accurate than typing the class name directly. For example, OWL-ViT can not detect \textit{tram} correctly if only \textit{train} is entered as input. To solve this problem, we borrowed the concept for Cityscapes proposed in SemiVL, as shown in Tab. \ref{tab:concepts}. For \textit{terrain}, we added a concept called \textit{roadside grass}.

\begin{table}[!ht]
\centering
\setlength{\tabcolsep}{4pt}
\caption{Class concepts on Cityscapes. By default, it will follow the concept suggested by SemiVL \cite{hoyer2023semivl}. The concept marked with a dagger ($\dagger$) is a new concept we added.}
\label{tab:concepts}
\begin{tabular}{@{}clc@{}}
\toprule
Category & Class & Concepts \\
\midrule
nature & vegetation & vegetation, tree, hedge \\
nature & terrain & terrain, grass, soil, sand, roadside grass$^\dagger$ \\
vehicle & car & car, jeep, SUV, van \\
vehicle & truck & truck, box truck, pickup truck, truck trailer \\
vehicle & bus & bus \\
vehicle & train & train, tram \\
vehicle & motorcycle & motorcycle, moped, scooter \\
vehicle & bicycle & bicycle \\
\bottomrule
\end{tabular}
\end{table}

\section{Representation Analysis}
\label{sec:rep_analysis}

\begin{figure}[!ht]
  \centering
  \includegraphics[width=\linewidth]{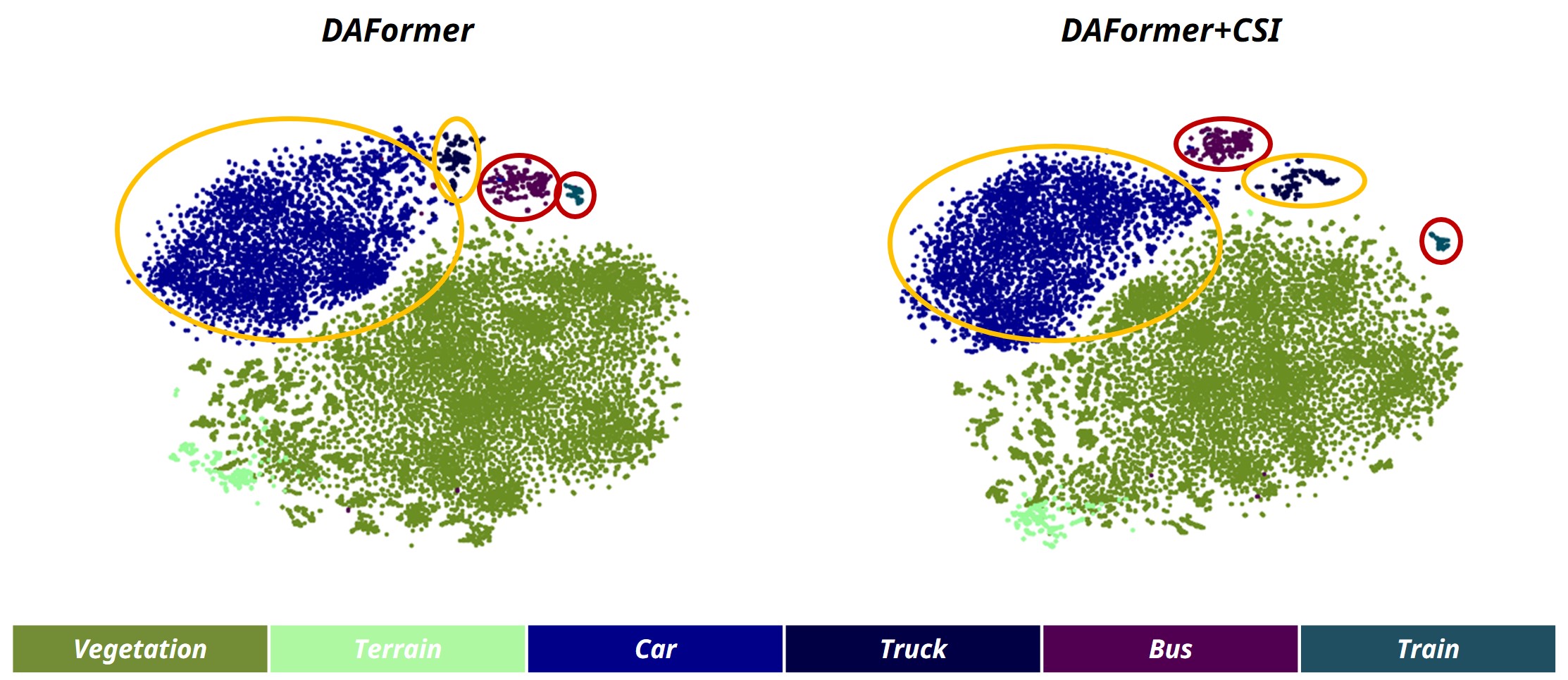}
  \caption{T-SNE embedding of the last bottleneck feature of the model for the Cityscapes validation set. The model is trained on Synthia-to-Cityscapes. The red and yellow circles show that \textit{bus-train} and \textit{car-truck} are better distinguished in CSI than in DAFormer.}
  \label{fig:t-sne}
\end{figure}

We analyze how well CSI generates representation compared to a base model. We believe this is a good way to check if the model has the new semantics well. Fig. \ref{fig:t-sne} visualizes the last bottleneck feature of the model trained on Synthia-to-Cityscapes for the Cityscapes validation set. We select the classes \textit{vegetation}, \textit{terrain}, \textit{car}, \textit{truck} of the coarse-to-fine taxonomy and \textit{train} of the open taxonomy for visualization. In addition, we add \textit{bus} to analyze using T-SNE because \textit{bus} is often mapped and relabeled as \textit{train}. The results were colored using ground truth. The red and yellow circles show that DAFormer represents \textit{bus-train} or \textit{car-truck} closely. On the other hand, CSI has a clear distinction between clusters of each class. Through this, it can be seen that CSI is clearly distinguishing the new class from the common classes. However, there is still a limitation that \textit{vegetation} and \textit{terrain} are less distinguished. This seems to be due to the low performance of \textit{terrain} in Synthia-to-Cityscapes.

\section{Additional Qualitative Analysis}

\label{sec:qual_analysis}
\begin{figure}[!ht]
  \centering
  \includegraphics[width=\textwidth]{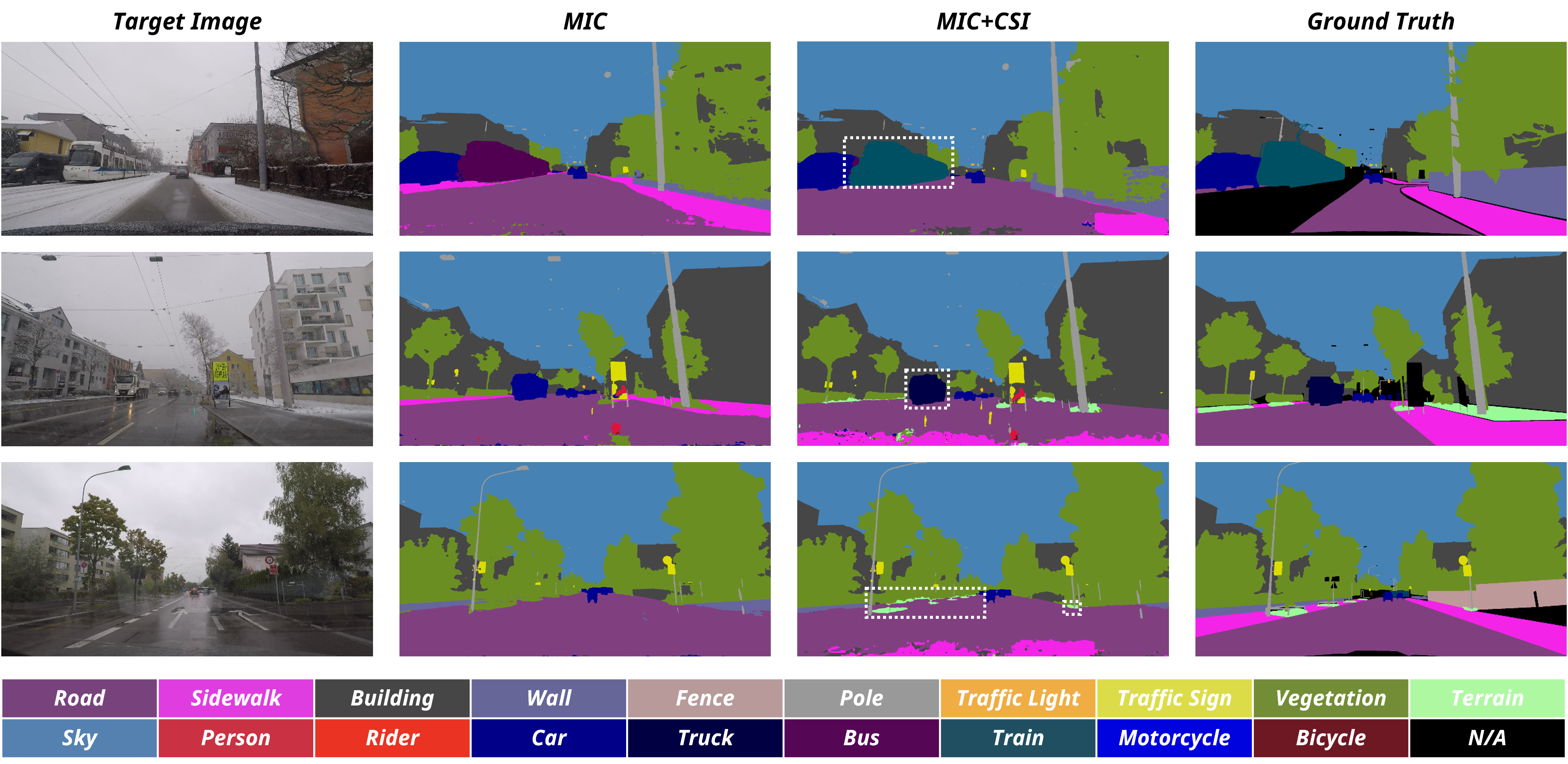}
  \caption{Example predictions on Synthia-to-ACDC}
  \label{fig:qual_acdc}
\end{figure}

We also tested our proposed method, CSI, on Synthia-to-ACDC. From this experiment, we can see that our method works well on different kinds of target domains. As shown in Fig. \ref{fig:qual_acdc}, ACDC is a harsher environment for segmentation than Cityscapes. Nevertheless, CSI can learn the semantics of the new classes from the zero-shot model and segment them well. Our proposed method still holds a lot of promise for further development like adapting from the source domain with a much smaller number of classes to the target domain.

\section{Discussion}
\label{sec:discussion}

Our CSI leverages vision language models (CLIP or OWL-ViT) which are pre-trained on large-scale datasets. This raises the question of whether our method is still unsupervised domain adaptation. We would like to discuss this question based on previous UDA studies.

Commonly, the encoder of the semantic segmentation model is initialized using the weight of the encoder pre-trained on a large-scale dataset. One of the representative previous UDA studies, DAFormer, uses an encoder pre-trained on ImageNet for classification. In DAFormer, not only initializing the model with an ImageNet encoder but also using the frozen encoder (pre-trained on ImageNet containing real-world images) as useful guidance for the thing class. They observed that the model segments some classes well at the beginning of training, but then forgets them after a few hundred training steps. To address this, they made sure that the model was not overfitted to the source domain through feature distance loss using a frozen image encoder. This method was particularly effective in segmenting \textit{train} and \textit{bus} classes well. Many recent UDA studies, including MIC and HRDA, have been studied based on DAFomer. Therefore, it can be viewed as common practice to use a large-scale dataset containing a real-world image in UDA studies. of course, we agree that our method relies more on the knowledge of large-scale datasets than previous studies, given that OWL-ViT pre-trained its detection box on the LVIS \cite{gupta2019lvis} dataset. However, as shown in Tab. \ref{tab:intersection}, there are no labels for stuff classes like terrain in the large-scale dataset. Nevertheless, CSI achieves 70-80\% of the performance of traditional UDA on GTA-to-Cityscapes for \textit{terrain}.

\begin{table}
\centering
\caption{Brief intersection between the classes in the target domain and the detected class lists used in the OWL-ViT and UDA model encoders.}
\label{tab:intersection}
\centering\resizebox{\textwidth}{!}{
\begin{tabular}{@{}ccc@{}}
\toprule
Class & ImageNet-1K & LVIS \\
\midrule
terrain & - & - \\
\makecell[t]{truck} & \makecell[t]{fire engine, fire truck \\ pickup, pickup truck, \\ tow truck, tow car, wrecker, \\ trailer truck, tractor trailer, \\  garbage truck, dustcart, \\ trucking rig, rig \\ articulated lorry, semi} & \makecell[t]{garbage truck, pickup truck, \\ tow truck, trailer truck, truck} \\
\makecell[t]{train} & \makecell[t]{bullet train, tram, tramcar, \\ streetcar, trolley, trolley car, \\ electric locomotive, steam locomotive} & \makecell[t]{bullet train, railcar (part of a train), \\ passenger car (part of a train), \\ train (railroad vehicle)}\\
\bottomrule
\end{tabular}}
\end{table}

\end{document}